\documentclass[runningheads]{llncs}

\usepackage[misc]{ifsym}
\usepackage{graphicx}
\usepackage[ruled,vlined]{algorithm2e}
\usepackage{xcolor}
\usepackage{caption}
\usepackage{booktabs}
\usepackage{subcaption}
\usepackage{hyperref}
\usepackage{fancyhdr}

\fancypagestyle{firststyle}
{
  \fancyhf{}

  \fancyhead[L]{\footnotesize Published as a conference paper at ECMLPKDD 2020}
}

\begin{document}
\title{Topological Insights into Sparse Neural Networks}
\titlerunning{Topological Insights into Sparse Neural Networks }
%
\author{Shiwei Liu[\Letter]\inst{1} \and
Tim Van der Lee\inst{1} \and
Anil Yaman\inst{1,4} \and
Zahra Atashgahi\inst{1,2} \and
Davide Ferraro\inst{3} \and
Ghada Sokar\inst{1} \and
Mykola Pechenizkiy\inst{1} \and
Decebal Constantin Mocanu\inst{1,2}}

\authorrunning{S. Liu et al.}

\institute{
Department of Mathematics and Computer Science, Eindhoven University of Technology, 5600 MB Eindhoven, the Netherlands
\email{\{s.liu3,t.lee,a.yaman,z.atashgahi,g.a.z.n.sokar,m.pechenizkiy,\}@tue.nl}
\and
Faculty of Electrical Engineering, Mathematics and Computer Science, University of Twente, Enschede, 7522NB, the Netherlands\\
\email{d.c.mocanu@utwente.nl}\\
 \and
The Biorobotics Institute, Scuola Superiore Sant’Anna, Italy\\
\email{davide.ferraro@santannapisa.it}\\
\and 
Korea Advanced Institute of Science and Technology, Daejeon, 34141, Republic of Korea
\email{}\\
}
\maketitle           

\thispagestyle{firststyle}

\begin{abstract}
Sparse neural networks are effective approaches to reduce the resource requirements for the deployment of deep neural networks. Recently, the concept of adaptive sparse connectivity, has emerged to allow training sparse neural networks from scratch by optimizing the sparse structure during training. However, comparing different sparse topologies and determining how sparse topologies evolve during training, especially for the situation in which the sparse structure optimization is involved, remain as challenging open questions. This comparison becomes increasingly complex as the number of possible topological comparisons increases exponentially with the size of networks. In this work, we introduce an approach to understand and compare sparse neural network topologies from the perspective of graph theory. We first propose Neural Network Sparse Topology Distance (NNSTD) to measure the distance between different sparse neural networks. Further, we demonstrate that sparse neural networks can outperform over-parameterized models in terms of performance, even without any further structure optimization. To the end, we also show that adaptive sparse connectivity can always unveil a plenitude of sparse sub-networks with very different topologies which outperform the dense model, by quantifying and comparing their topological evolutionary processes. The latter findings complement the \textit{Lottery Ticket Hypothesis} by showing that there is a much more efficient and robust way to find ``winning tickets". Altogether, our results start enabling a better theoretical understanding of sparse neural networks, and demonstrate the utility of using graph theory to analyze them.


\end{abstract}
\section{Introduction}
Deep neural networks have led to promising breakthroughs in various applications. While the performance of deep neural networks improving, the size of these usually over-parameterized models has been tremendously increasing. The training and deploying cost of the state-of-art models, especially pre-trained models like BERT \cite{devlin2018bert}, is very large. 

Sparse neural networks are an effective approach to address these challenges. Discovering a small sparse and well-performing sub-network of a dense network can significantly reduce the parameters count (e.g. memory efficiency), along with the floating-point operations. Over the past decade, many works have been proposed to obtain sparse neural networks, including but not limited to magnitude pruning \cite{han2015learning,guo2016dynamic}, Bayesian statistics \cite{molchanov2017variational,louizos2017bayesian}, $l_0$ and $l_1$ regularization \cite{louizos2017learning}, reinforcement learning \cite{lin2017runtime}. Given a pre-trained model, these methods can efficiently discover a sparse sub-network with competitive performance. While some works aim to provide analysis of sparse neural networks \cite{frankle2018lottery,zhou2019deconstructing,gale2019state,liu2018rethinking}, they mainly focus on how to empirically improve training performance or to what extent the initialization and the final sparse structure contribute to the performance. Sparsity (the proportion of neural network weights that are zero-valued) inducing techniques essentially uncover the optimal sparse topologies (sub-networks) that, once initialized in a right way, can reach a similar predictive performance with dense networks as shown by the \textit{Lottery Ticket Hypothesis} \cite{frankle2018lottery}. Such sub-networks are named ``winning lottery tickets" and can be obtained from pre-trained dense models, which makes them inefficient during the training phase. 

Recently, many works have emerged to achieve both, training efficiency and inference efficiency, based on adaptive sparse connectivity \cite{mocanu2018scalable,mostafa2019parameter,liu2019intrinsically,dettmers2019sparse,evci2019rigging}. Such networks are initialized with a sparse topology and can maintain a fixed sparsity level throughout training. Instead of only optimizing model parameters - weight values (continuous optimization problem), in this case, the sparse topology is also optimized (combinatorial optimization problem) during training according to some criteria in order to fit the data distribution. In \cite{evci2019rigging}, it is shown that such metaheuristics approaches always lead to very-well performing sparse topologies, even if they are based on a random process, without the need of a pre-trained model and a \textit{lucky} initialization as done in \cite{frankle2018lottery}. While it has been shown empirically that both approaches, i.e. winning lottery tickets and adaptive sparse connectivity, find very well-performing sparse topologies, we are generally lacking their understanding. Questions such as: \textit{How different are these well-performing sparse topologies?}, \textit{Can very different sparse topologies lead to the same performance?}, \textit{Are there many local sparse topological optima which can offer sufficient performance (similar in a way with the local optima of the weights continuous optimization problem)?}, are still unanswered. 

In this paper, we are studying these questions in order to start enabling a better theoretical understanding of sparse neural networks and to unveil high gain future research directions. Concretely, our contributions are:
\begin{itemize}
    \item We propose the first metric which can measure the distance between two sparse neural networks topologies\footnote{Our code is available at\\ \url{https://github.com/Shiweiliuiiiiiii/Sparse_Topology_Distance}}, and we name it Neural Network Sparse Topology Distance (NNSTD). For this, we treat the sparse network as a large \textit{neural graph}. In NNTSD, we take inspiration  from graph theory and Graph Edit Distance (GED) \cite{6313167} which cannot be applied directly due to the fact that two different neural graphs may represent very similar networks since hidden neurons are interchangeable \cite{li2015convergent}.
    \item Using NNSTD, we demonstrate that there exist many very different well-performing sparse topologies which can achieve the same performance. \item In addition, with the help of our proposed distance metric, we confirm and complement the findings from \cite{evci2019rigging} by being able to quantify \textit{how different} are the sparse and, at the same time, similarly performing topologies obtained with adaptive sparse connectivity. This implicitly implies that there exist many local well-performing sparse  topological optima. 
\end{itemize}





\section{Related Work}
\subsection{Sparse Neural Networks}
\subsubsection{Sparse Neural Networks for Inference Efficiency.} 
Since being proposed, the motivation of sparse neural networks is to reduce the cost associated with the deployment of deep neural networks (inference efficiency) and to gain better generalization \cite{chauvin1989back,hassibi1993second,lecun1990optimal}. Up to now, a variety of methods have been proposed to obtain inference efficiency by compressing a dense network to a sparse one. Out of them, pruning is certainly the most effective one. A method which iteratively alternates pruning and retraining was introduced by Han et al. \cite{han2015learning}. This method can reduce the number of connections of AlexNet and VGG-16 on ImageNet by 9$\times$ to 13$\times$ without loss of accuracy. Further, Narang et al. \cite{narang2017exploring} applied pruning to recurrent neural networks while getting rid of the retraining process. At the same time, it is shown in \cite{zhu2017prune} that, with the same number of parameters, the pruned models (large-sparse) have better generalization ability than the small-dense models. A grow-and-prune (GP) training was proposed in \cite{dai2018grow}. The network growth phase slightly improves the performance. While unstructured sparse neural networks achieve better performance, it is difficult to be applied into parallel processors, since the limited support for sparse operations. Compared with fine-grained pruning, coarse-grained (filter/channel) pruning is more desirable to the practical application as it is more amenable for hardware acceleration \cite{he2018soft,he2019filter}. 

\subsubsection{Sparse Neural Networks for Training Efficiency} 
Recently, more and more works attempt to get memory and computational efficiency for the training phase. This can be naturally achieved by training sparse neural networks directly. However, while training them with a fixed sparse topology can lead to good performance \cite{gotMocanuEcml2016}, it is hard to find an optimal sparse topology to fit the data distribution before training. This problems was addressed by introducing the adaptive sparse connectivity concept through its first instantiation, the Sparse Evolutionary Training (SET) algorithm \cite{dcmocanuphd,mocanu2018scalable}. SET is a straightforward strategy that starts from random sparse networks and can achieve good performance based on magnitude weights pruning and regrowing after each training epoch. Further, Dynamic Sparse Reparameterization (DSR) \cite{mostafa2019parameter} introduced across-layer weights redistribution to allocate more weights to the layer that contributes more to the loss decrease. By utilizing the momentum information to guide the weights regrowth and across-layer redistribution, Sparse Momentum \cite{dettmers2019sparse} can improve the classification accuracy for various models. However, the performance improvement is at the cost of updating and storing the momentum of every individual weight of the model. Very recently, instead of using the momentum, The Rigged Lottery \cite{evci2019rigging} grows the zero-weights with the highest magnitude gradients to eliminate the extra floating point operations required by Sparse Momentum. Liu et al. \cite{liu2019intrinsically} trained intrinsically sparse recurrent neural networks (RNNs) that can achieve usually better performance than dense models. Lee et al \cite{lee2018snip} introduced single-shot network pruning (SNIP) that can discover a sparse network before training based on a connection sensitivity criterion. Trained in the standard way, the sparse pruned network can have good performance. Instead of using connection sensitivity, GraSP \cite{wang2020picking} prunes connections whose removal causes the least decrease in the gradient norm, resulting in better performance than SNIP in the extreme sparsity situation.

\subsubsection{Interpretation and Analysis of Sparse Neural Networks}
Some works are aiming to interpret and analyze sparse neural networks. Frankle \& Carbin \cite{frankle2018lottery} proposed the \textit{Lottery Ticket Hypothesis} and shown that the dense structure contains sparse sub-networks that are able to reach the same accuracy when they are trained with the same initialization. Zhou et al. \cite{zhou2019deconstructing} further claimed that the sign of the ``lucky'' initialization is the key to guarantee the good performance of "winning lottery tickets". Liu et al. \cite{liu2018rethinking} reevaluated the value of network pruning techniques. They showed that training a small pruned model from scratch can reach the same or even better performance than conventional network pruning and for small pruned models, the pruned architecture itself is more crucial to the learned weights. Moreover, magnitude pruning \cite{zhu2017prune} can achieve better performance than $l_0$ regularization \cite{louizos2017learning} and variational dropout \cite{molchanov2017variational} in terms of large-scale tasks \cite{gale2019state}. 

\subsection{Sparse Evolutionary Training}

Sparse Evolutionary Training (SET) \cite{mocanu2018scalable} is an effective algorithm that allows training sparse neural networks from scratch with a fixed number of parameters. Instead of starting from a highly over-parameterized dense network, the network topology is initialized as a sparse Erd{\H{o}}s-R{\'e}nyi graph \cite{gilbert1959random}, a graph where each edge is chosen randomly with a fixed probability, independently from every other edge. Given that the random initialization may not always guarantee good performance, adaptive sparse connectivity is utilized to optimize the sparse topology during training. Concretely, a fraction $\zeta$ of the connections with the smallest magnitude are pruned and an equal number of novel connections are re-grown after each training epoch. This adaptive sparse connectivity (pruning-and-regrowing) technique is capable of guaranteeing a constant sparsity level during the whole learning process and also improving the generalization ability. More precisely, at the beginning of the training, the connection (${W}^k_{ij}$) between neuron $h_j^{k-1}$ and $h_i^k$ exists with the probability:  
\begin{equation}
 p({W}^k_{ij})=\min(\frac{\epsilon(n^k+n^{k-1})}{n^kn^{k-1}}, 1)
 \label{Eq:probtopology}
\end{equation}
where $n^{k}, n^{k-1}$ are the number of neurons of layer $h^{k}$ and $h^{k-1}$, respectively;
$\epsilon$ is a parameter determining the sparsity level. The smaller $\epsilon$ is, the more sparse the network is. By doing this, the sparsity level of layer $k$ is given by $1-\frac{\epsilon(n^k+n^{k-1})}{n^kn^{k-1}}$.
The connections between the two layers are collected in a sparse weight matrix $\textbf{W}^k \in \textbf{R}^{n^{k-1}\times{n^{k}}}$. Compared with fully-connected layers whose number of connections is $n^{k}n^{k-1}$ , the SET sparse layers only have $n^W = \| \textbf{W}^k \|_0 = \epsilon(n^{k}+n^{k-1})$ connections which can significantly alleviate the pressure of the expensive memory footprint. Among all possible adaptive sparse connectivity techniques, in this paper, we make use of SET due to two reasons: (1) its natural simplicity and computational efficiency, and (2) the fact that the re-grown process of new connections is purely random favoring in this way an unbiased study of the evolved sparse topologies.

\section{Neural Network Sparse Topology Distance}
In this section, we introduce our proposed method, NNSTD, to measure the topological distance between two sparse neural networks. The sparse topology locution used in this paper refers to the graph underlying a sparsely connected neural network in which each neuron represents a vertex in this graph and each existing connection (weight) represents an edge in the graph. Existing metrics to measure the distance between two graphs are not always applicable to artificial neural network topologies. The main difficulty is that two different graph topologies may represent similar neural networks since hidden neurons are interchangeable. All graph similarity metrics consider either labeled or unlabeled nodes to compute the similarity. With neural networks, input and output layers are labeled (each of their neurons corresponds to a concrete data feature or class, respectively), whereas hidden layers are unlabelled. In particular, we take multilayer perceptron networks (MLP) as the default. 

The inspiration comes from Graph Edit Distance (GED) ~\cite{6313167}, a well-known graph distance metric. Considering two graphs $g_1$ and $g_2$, it measures the minimum cost required to transform $g_1$ into a graph isomorphic to $g_2$. Formally the graph edit distance is calculated as follows.
\begin{equation}
    GED(g_1,g_2) = \min_{p \in P(g_1,g_2)} c(p)
\end{equation}
where $p$ represents a sequence of transformation from $g_1$ into a graph isomorphic to $g_2$, and $c$ represents the total cost of such transformation. $P$ represents all possible transformations. This large panel of possibilities makes computing the GED a NP-hard problem when a subset of the nodes in the graphs are unlabeled (e.g. hidden neurons are interchangeable).

\begin{figure}[!t]
\centering
\includegraphics[width=\textwidth]{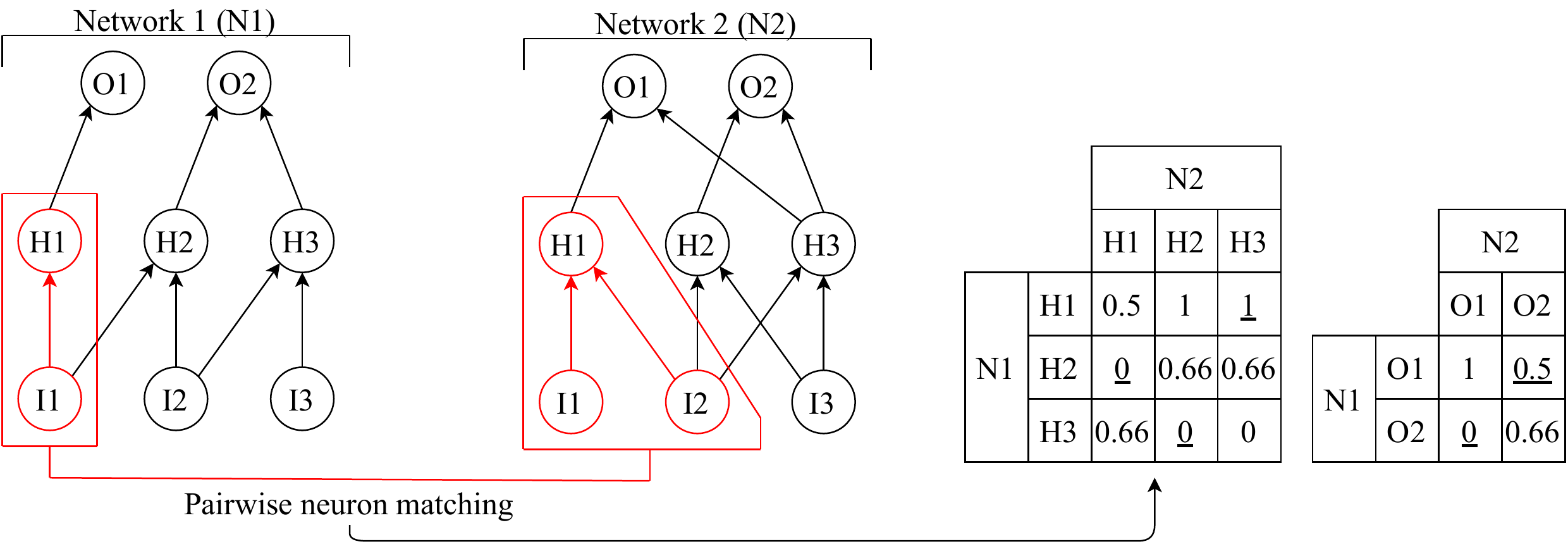}
\caption{NNSTD metric illustration.}
\label{fig:method}
\end{figure}

\begin{algorithm}[ht!]
\small
\caption{Neural Network Sparse Topology Distance} \label{algo:method}
\SetAlgoLined
\SetKwFunction{FGed}{NED}
\SetKwFunction{FComp}{CompareLayers}
\SetKwFunction{FMain}{CompareNetworks}
\SetKwProg{Fn}{Function}{:}{}
\Fn{\FGed{$G1,G2$}}{
    \textbf{return} $\frac{\vert(G1\setminus G2)\cup(G2\setminus G1)\vert}{\vert(G1\setminus G2)\cup(G2\setminus G1)\cup(G1\cap G2)\vert}$\;
}

\Fn{\FComp{$L1,L2$}}{
    $\texttt{NNSTDmatrix}$\;
    \For{neuron $n1$ in $L1$}{
        \For{neuron $n2$ in $L2$}{
            $G1 = input\_neurons\_set\_of(n1)$\;
            $G2 = input\_neurons\_set\_of(n2)$\;
            $\texttt{NNSTDmatrix}[(n1,n2)] = NED(G1,G2)$\;
        }
    }
    $\texttt{neuron\_assignment}, \texttt{normalized\_cost} = solve(\texttt{NNSTDmatrix})$\;
    \textbf{return} \texttt{neuron\_assignment}, \texttt{normalized\_cost}/size$(\L2)$\;
}

\Fn{\FMain{$N1,N2$}}{
    $\texttt{neuron\_assignment} = \texttt{Identity}$\;
    $\texttt{costs} = 0$\;
    \For{layer $l$ in $[1,L]$}{
        $\texttt{neuron\_assignment}, \texttt{normalized\_cost} = \FComp{N1[l],N2[l]}$\;
        $\texttt{reorder}(N2[l],\texttt{neuron\_assignment})$\;
        $\texttt{costs} += \texttt{normalized\_cost}$\;
    }
    \textbf{return} $\texttt{costs}/L$\;
} 
\end{algorithm}
The proposed NNSTD metric is presented in Algorithm~\ref{algo:method} and discussed next. A graphical example is also provided in Figure~\ref{fig:method}. As an example, two neural networks $(N1, N2)$ are considered. For each hidden neuron $n$, a tree graph is constructed based on all direct inputs to this neuron, and these input neurons are collected in a set $g$. Per layer, for all possible pairs of neurons between the two networks, the Normalized Edit Distance (NED) is calculated between their input neurons, as defined in the second line of Algorithm~\ref{algo:method}. NED takes the value 1 if the two compared neurons have no input neurons in common, and 0 if they have the exact same neurons as input. To reduce the complexity of the search space, we take a greedy approach, and for any current layer we consider that the neurons of the previous layer are labeled (as they have been matched already by the proposed distance metric when the previous layer was under scrutiny), and that adding or deleting inputs have the same cost. For instance, for the neurons compared in Figure~\ref{fig:method}, one input neuron is shared out of two different inputs considered, thus the distance between them is $NED(N1:H1, N2:H1)=0.5$.
The NNSTD matrix is solved using the Hungarian method to find the neuron (credit) assignment problem which minimizes the total cost, presented in underlined Figure~\ref{fig:method}. The aggregated costs divided by the size of $L2$ gives the distance between the first layer of $N1$ and $N2$. To compare the next layer using the same method, the current layer must be fixed. Therefore the assignment solving the NNSTD matrix is saved to reorder the first layer of $N2$. To the end, an NNSTD value of 0 between two sparse layers (or two sparse networks) shows that the two layers are exactly the same, while a value of 1 (maximum possible) shows that the two layers are completely different.


\section{Experimental Results}

In this section, we study the performance of the proposed NNSTD metric and the sparse neural network properties on two datasets, Fashion-MNIST \cite{xiao2017fashion} and CIFAR-10 \cite{krizhevsky2009learning}, in a step-wise fashion. We begin in Section \ref{para:performance} by showing that sparse neural networks can match the performance of the fully-connected counterpart, even without topology optimization. Next, in Section \ref{Topology Distance of Sparse Neural Networks} we first validate NNSTD and then we apply it to show that adaptive sparse connectivity can find many well-performing very different sub-networks. Finally, we verify that adaptive sparse connectivity indeed optimizes the sparse topology during training in Section \ref{para:verifyASC}.

\subsection{Experimental Setup}
For the sake of simplicity, the models we use are MLPs with SReLU activation function \cite{jin2016deep} as it has been shown to provide better performance for SET-MLP \cite{mocanu2018scalable}. For both datasets, we use 20\% of the training data as the validation set and the test accuracy is computed with the model that achieves the highest validation accuracy during training. 

For Fashion-MNIST, we choose a three-layer MLP as our basic model, containing 784 hidden neurons in each layer. We set the batch size to 128. The optimizer is stochastic gradient descent (SGD) with Nesterov momentum. We train these sparse models for 200 epochs with a learning rate of 0.01, Nesterov momentum of 0.9. And the weight decay is 1e-6. 
 
The network used for CIFAR-10 consists of two hidden layers with 1000 hidden neurons. We use standard data augmentations (horizontal flip, random rotate, and random crop with reflective padding). We set the batch size to 128. We train the sparse models for 1000 epochs using a learning rate of 0.01, stochastic gradient
descent with Nesterov momentum of $\alpha$ = 0.9. And we use a weight decay of 1e-6. 


\subsection{The Performance of Sparse Neural Networks}
\label{para:performance}
We first verify that random initialized sparse neural networks are able to reach a competitive performance with the dense networks, even without any further topology optimization. 

\begin{figure}[!ht]
\centering
\begin{subfigure}[b]{0.45\textwidth}
    \includegraphics[width=\textwidth]{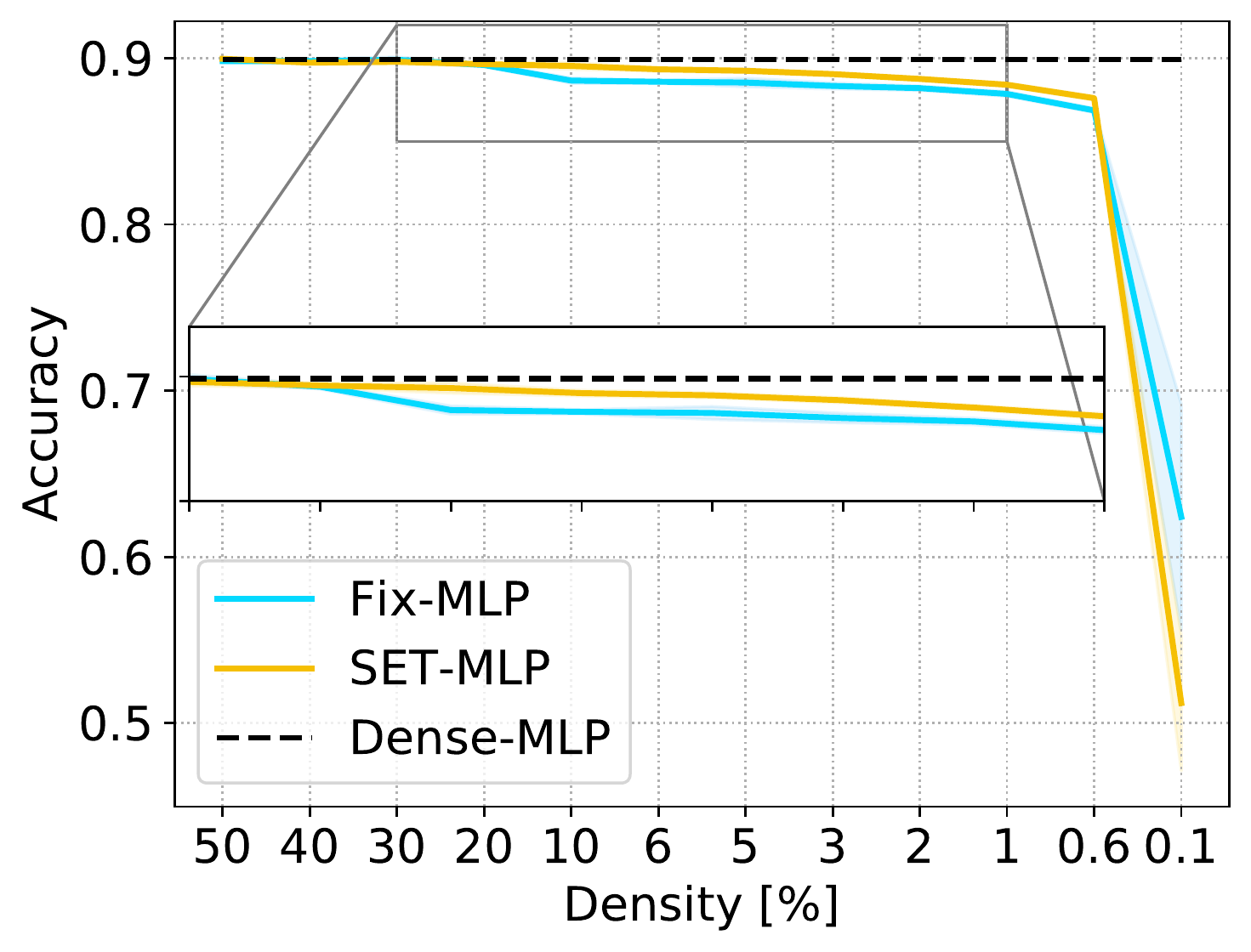}
    \caption{Test accuracy with three-layer MLPs on Fashion-MNIST.}
    \label{fig:tc1}
\end{subfigure}
~
\begin{subfigure}[b]{0.463\textwidth}
    \includegraphics[width=\textwidth]{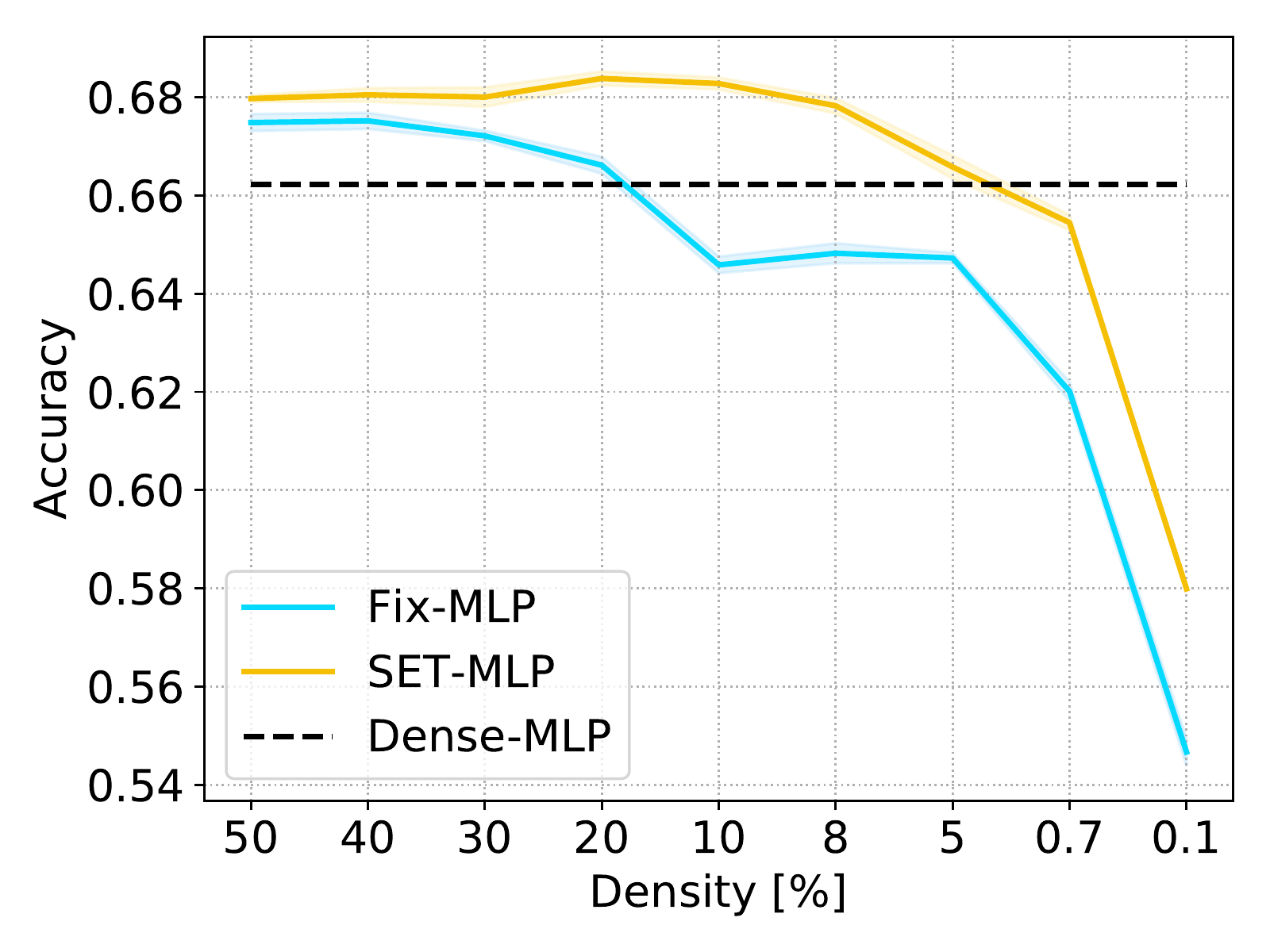}
    \caption{Test accuracy with two-layer MLPs on CIFAR-10.}
    \label{fig:tc2}
\end{subfigure}
~
\caption{Test accuracy of MLPs with various density levels. SET-MLP refers to the networks trained with adaptive sparse connectivity associated with SET. Fix-MLP refers to the networks trained without sparse topology optimization. The dashed lines represent the dense MLPs. Note that each line is the average of 8 trials and the standard deviation is very small.}
\label{fig:Accuracy_COM}
\end{figure}

For Fashion-MNIST, we train a group of sparse networks with density levels ($1 - sparsity$) in the space $\{0.1\%, 0.6\%, 1\%, 2\%, 3\%, 5\%, 6\%, 10\%, 20\%, 30\%, 40\%,\\ 50\%\}$. For each density level, we initialize two sparse networks with two different random seeds as root networks. For each root network, we generate a new network by randomly changing 1\% connections. We perform this generating operation 3 times to have 4 networks in total including the root network for each random seed. Every new network is generated from the previous generation. Thus, the number of networks for each density level is 8 and the total number of sparse networks of Fashion-MNIST is 96. We train these sparse networks without any sparse topology optimization for 200 epochs, named as Fix-MLP. To evaluate the effectiveness of sparse connectivity optimization, we also train the same networks with sparse connectivity optimization proposed in SET \cite{mocanu2018scalable} for 200 epochs, named as SET-MLP. The hyper-parameter of SET, pruning rate, is set to be 0.2. Besides this, we choose two fully-connected MLPs as the baseline. 

The experimental results are given in Figure \ref{fig:tc1}. We can see that, as long as the density level is bigger than 20\%, both Fix-MLP and SET-MLP can reach a similar accuracy with the dense MLP. While decreasing the density level decreases the performance of sparse networks gradually, sparse MLPs still reach the dense accuracy with only 0.6\% parameters. Compared with Fix-MLP, the networks trained with SET are able to achieve slightly better performance. 

For CIFAR-10, we train two-layer MLPs with various density levels located in the range $\{ 0.1\%, 0.7\%, 5\%, 8\%, 10\%, 20\%, 30\%, 40\%, 50\% \}$. We use the same strategy with Fashion-MNIST to generate 72 networks in total, 8 for each density level. All networks are trained with and without adaptive sparse connectivity for 900 epochs. The two-layer dense MLP is chosen as the baseline.

The results are illustrated in Figure \ref{fig:tc2}. We can observe that Fix-MLP consistently reaches the performance of the fully-connected counterpart when the percentage of parameters is larger than 20\%. It is more surprising that SET-MLP can significantly improve the accuracy with the help of adaptive sparse connectivity. With only 5\% parameters, SET-MLP can outperform the dense counterpart.

\subsection{Topological Distance between Sparse Neural Networks}
\label{Topology Distance of Sparse Neural Networks}
\subsubsection{Evaluation of Neural Network Sparse Topology Distance.}
\label{ESTS}

In this part, we evaluate our proposed NNSTD metric by measuring the initial topological distance between three-layer MLPs on Fashion-MNIST before training. We first measure the topology distance between networks with the same density. We initialize one sparse network with a density level of 0.6\%. Then, we generate 9 networks by iteratively changing 1\% of the connections from the previous generation step. By doing this, the density of these networks is the same, whereas the topologies vary a bit. Therefore, we expect that the topological distance of each generation from the root network should increase gradually as the generation adds up, but still to have a small upper bound. The distance measured by our method is illustrated in Figure \ref{fig:only5}. We can see that the result is consistently in line with our hypothesis. Starting with the value close to zero, the distance increases as the topological difference adds up, but the maximum distance is still very small, around 0.2.
\def\imagebox#1#2{\vtop to #1{\null\hbox{#2}\vfill}}
\begin{figure}[!ht]
\centering
 \hspace*{-1.2cm} 
\begin{subfigure}[b]{0.45\textwidth}
    \includegraphics[width=1.1\textwidth]{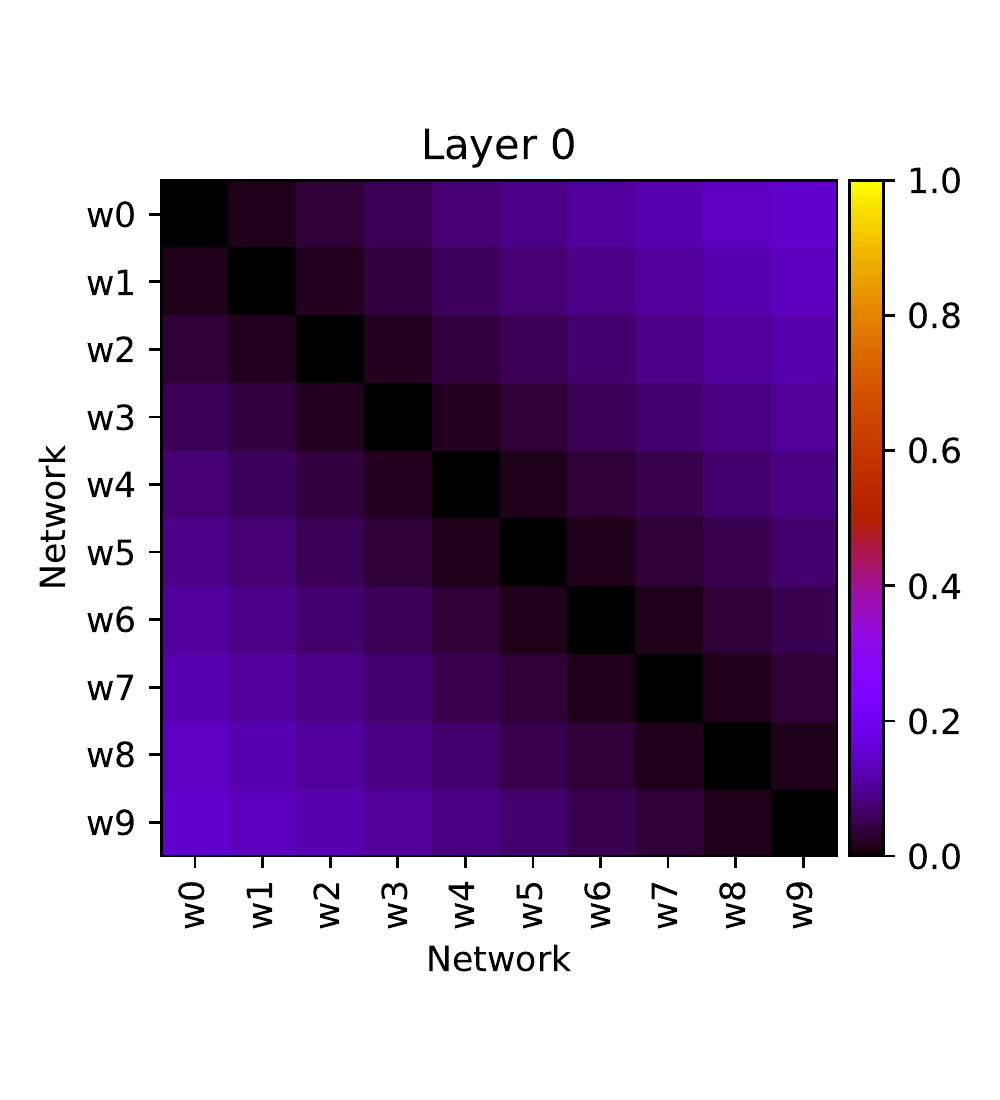}
    \caption{Same density level}
    \label{fig:only5}
\end{subfigure}
~
\hspace*{0.3cm} 
\begin{subfigure}[b]{0.46\textwidth}
    \imagebox{67mm}{\includegraphics[width=1.1\textwidth]{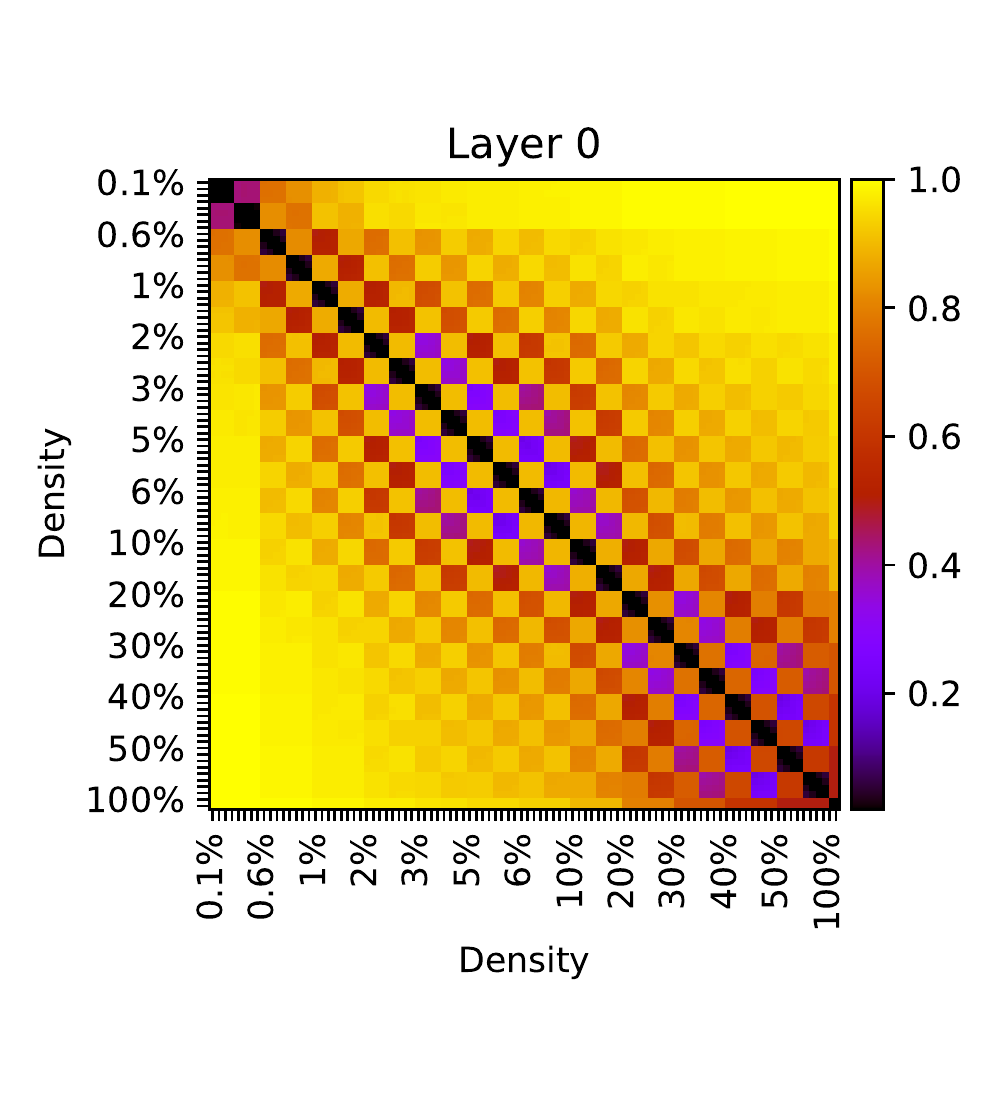}}
    \caption{Different density levels}
    \label{fig:s0m0}
\end{subfigure}
 \hspace*{-1.2cm} 
~ 

\caption{Evaluation of the proposed NNSTD metric. (a) refers to the sparse topology distance among 10 networks generated by randomly changing 1\% connections with the same density level of 0.6\%. $w_i (i=1,2,...,9)$ represents these gradually changed networks. (b) represents the sparse topology distance among networks generated with different density levels.}
 \vskip -0.3cm
\label{fig:evaluation}
\end{figure}

Further, we also evaluate NNSTD on sparse networks with different density levels. We use the same 96 sparse and two dense networks generated in Section \ref{para:performance}. Their performance is given in Figure \ref{fig:tc1}. Concretely, for each density level, we choose 8 networks generated by two different random seeds. For each density level in the plot, the first four networks are generated with one random seed and the latter four networks are generated with another one. We hypothesize that distance among the networks with the same density should be different from the networks with different density. The distance among networks with different density can be very large, since the density varies over a large range, from 0.1\% to 100\%. Furthermore, the topological distance increases as the density difference increases, since more cost is required to match the difference between the number of connections. We show the initial topology distance in Figure \ref{fig:s0m0}. We can see that the distance among different density levels can be much larger than among the ones with the same density, up to 1. The more similar the density levels are, the more similar the topologies are. As expected, the distance between networks with the same density generated with different random seeds is very big. This makes sense as initialized with different random seeds, two sparse connectivities between two layers can be totally different. We only plot the distance of the first layer, as all layers are initialized in the same way. 
\subsubsection{Evolutionary Optimization Process Visualization.}
\label{evolution}
Herein, we visualize the evolutionary optimization process of the sparse topology learned by adaptive sparse connectivity associated with SET on Fashion-MNIST and CIFAR-10.

\begin{figure}[!ht]
\centering
\begin{subfigure}[b]{0.3\textwidth}
    \includegraphics[width=1.1\textwidth]{heatmapf_l0compare_paths_edit_opti_SET_e5__epoch000_set.pdf}
    \caption{Epoch 0}
    \label{fig:e5_0}
\end{subfigure}
~
\begin{subfigure}[b]{0.3\textwidth}
    \includegraphics[width=1.1\textwidth]{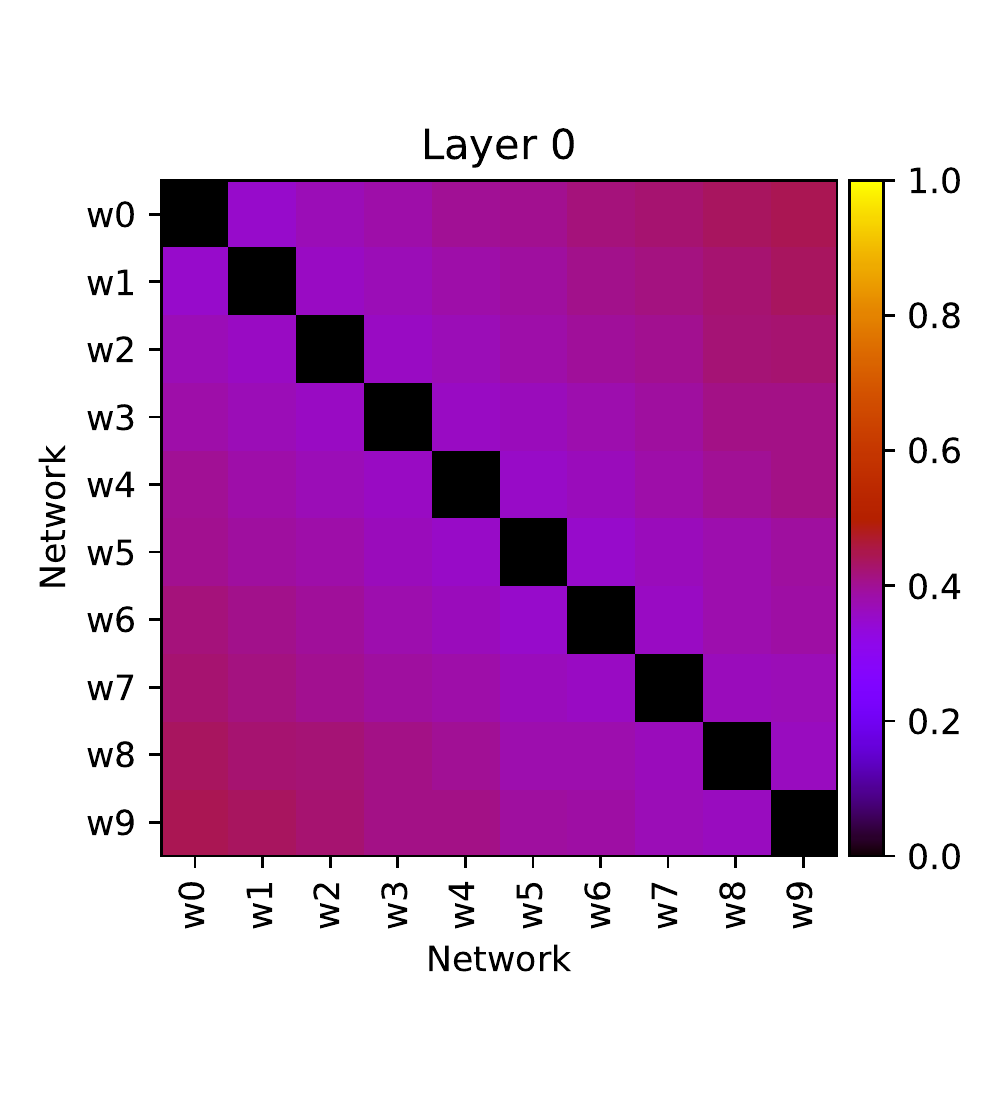}
    \caption{Epoch 10}
    \label{fig:e5_10}
\end{subfigure}
~
\begin{subfigure}[b]{0.3\textwidth}
    \includegraphics[width=1.1\textwidth]{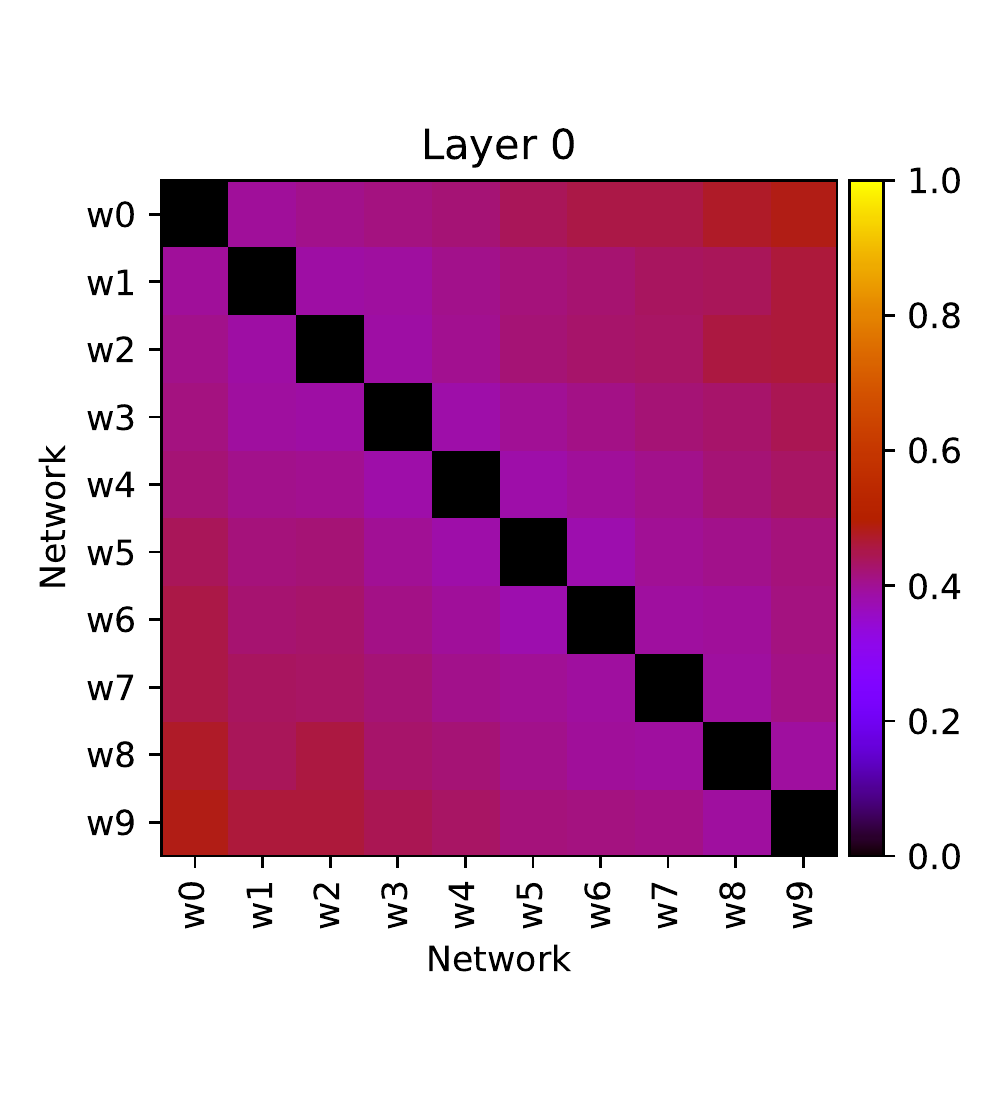}
    \caption{Epoch 30}
    \label{fig:e5_30}
\end{subfigure}
~
\begin{subfigure}[b]{0.3\textwidth}
    \includegraphics[width=1.1\textwidth]{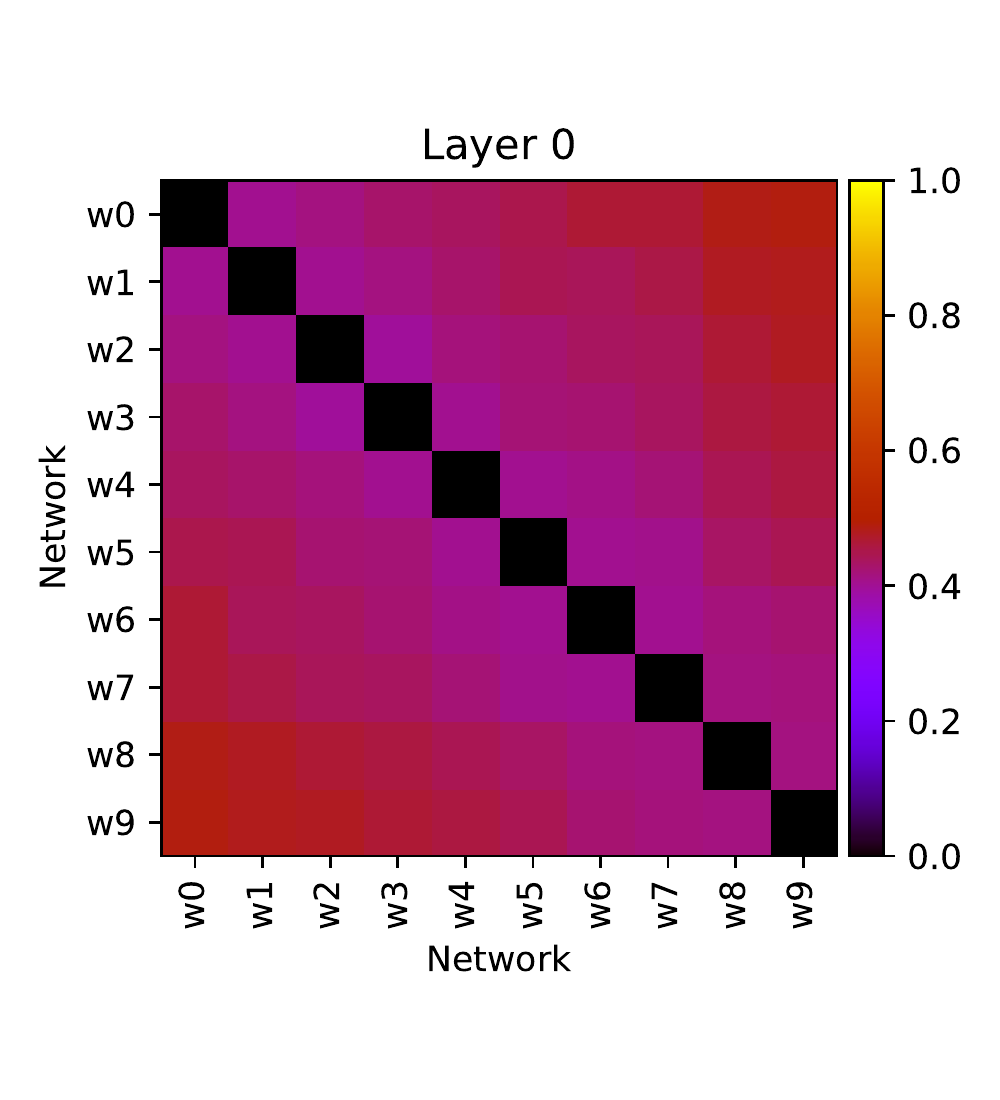}
    \caption{Epoch 50}
    \label{fig:e5_50}
\end{subfigure}
~
\begin{subfigure}[b]{0.3\textwidth}
    \includegraphics[width=1.1\textwidth]{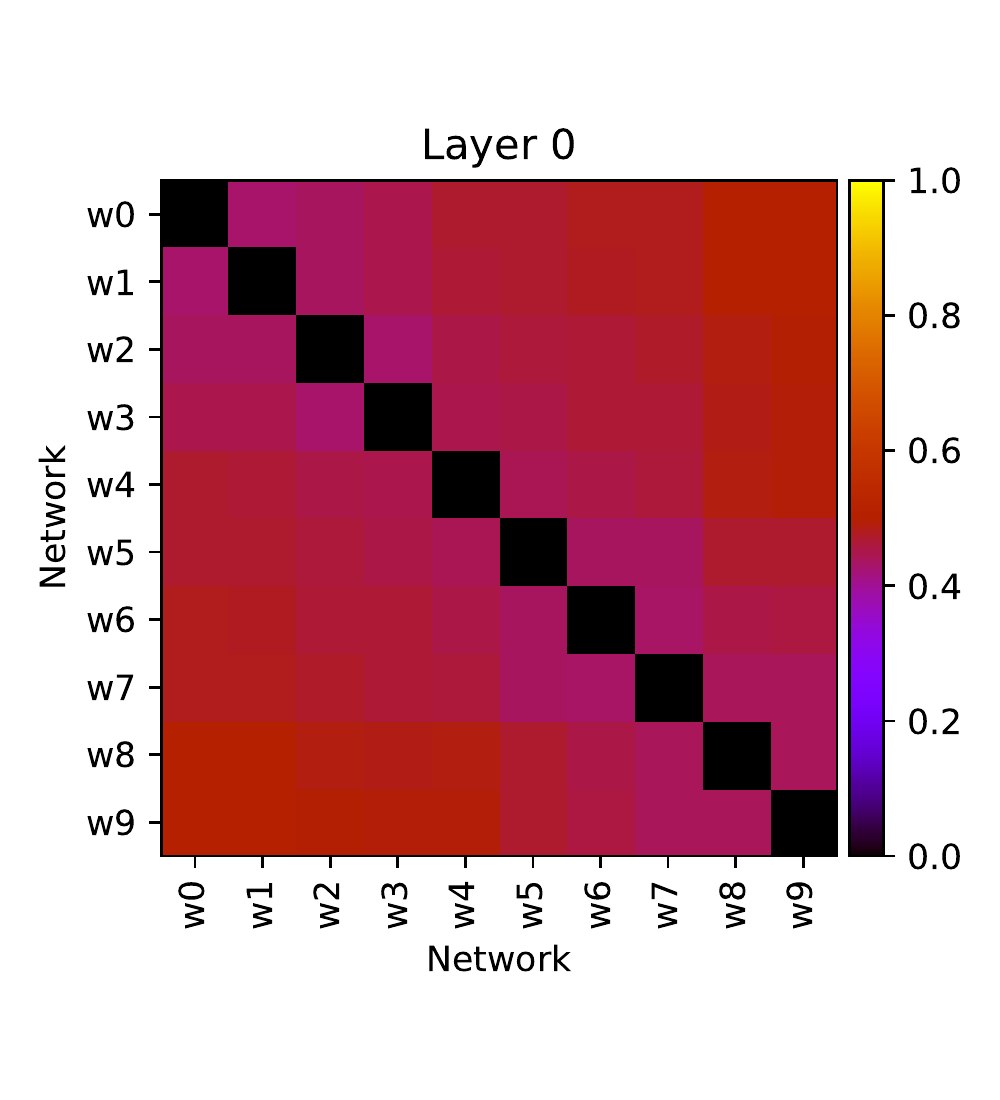}
    \caption{Epoch 100}
    \label{fig:e5_100}
\end{subfigure}
~
\begin{subfigure}[b]{0.3\textwidth}
    \includegraphics[width=1.1\textwidth]{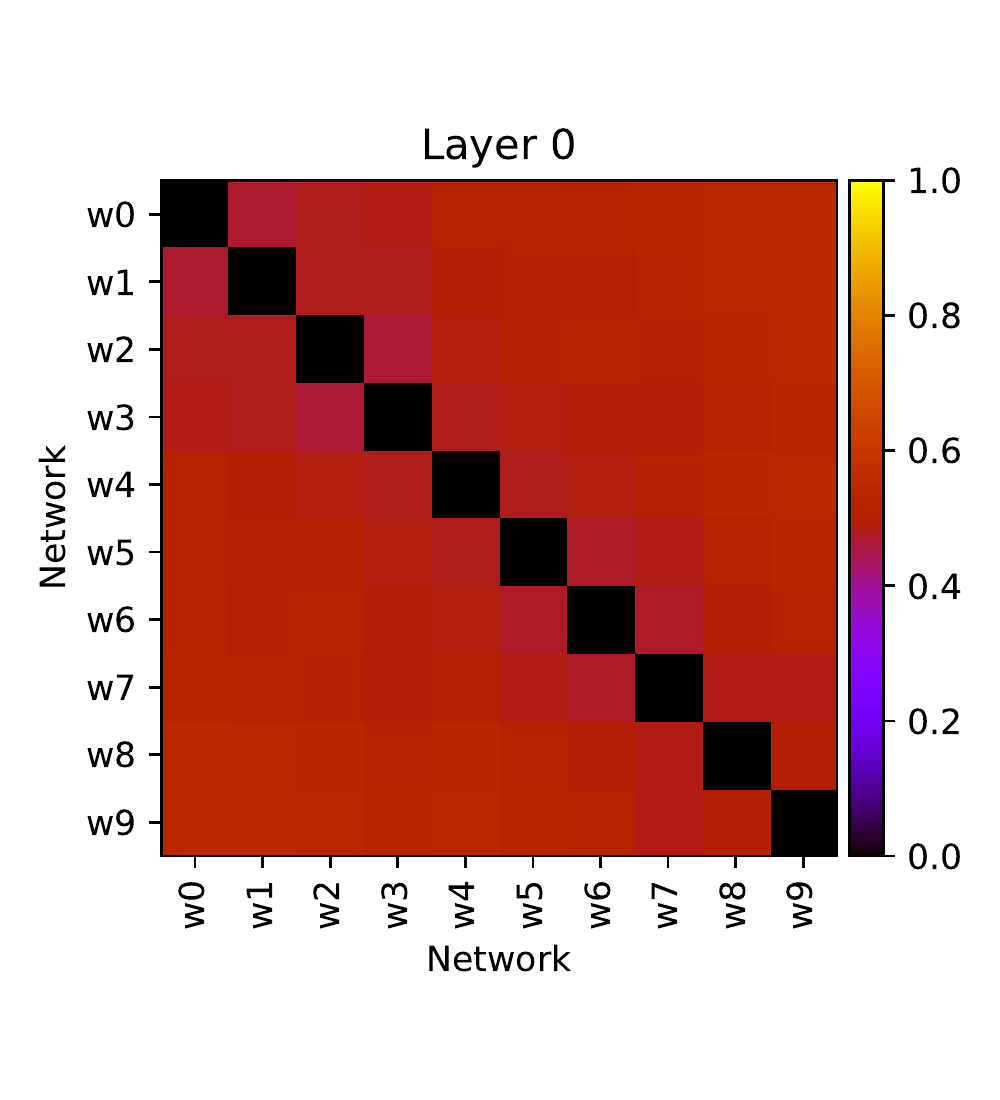}
    \caption{Epoch 190}
\end{subfigure}
\caption{Topological distance dynamics of 10 networks optimized by adaptive sparse connectivity with three-layer SET-MLP on Fashion-MNIST. The initial networks (epoch 0) have the same density level, with a tiny percentage (1\%) of topological difference with each other. $w_i (i=1,2,...,9)$ represents different networks.}
\label{fig:e5}
\end{figure}

\begin{figure}[!ht]
\centering
 \hspace*{-1.5cm} 
\begin{subfigure}[b]{0.45\textwidth}
    \includegraphics[width=1.0\textwidth]{heatmapf_l0compare_paths_edit_opti__s0s1_w0w1w2w3_l0.pdf}
    \caption{The initial distance.}
    \label{fig:evolve_96_original}
\end{subfigure}
~
\hspace*{0.8cm} 
\begin{subfigure}[b]{0.45\textwidth}
    \includegraphics[width=1.0\textwidth]{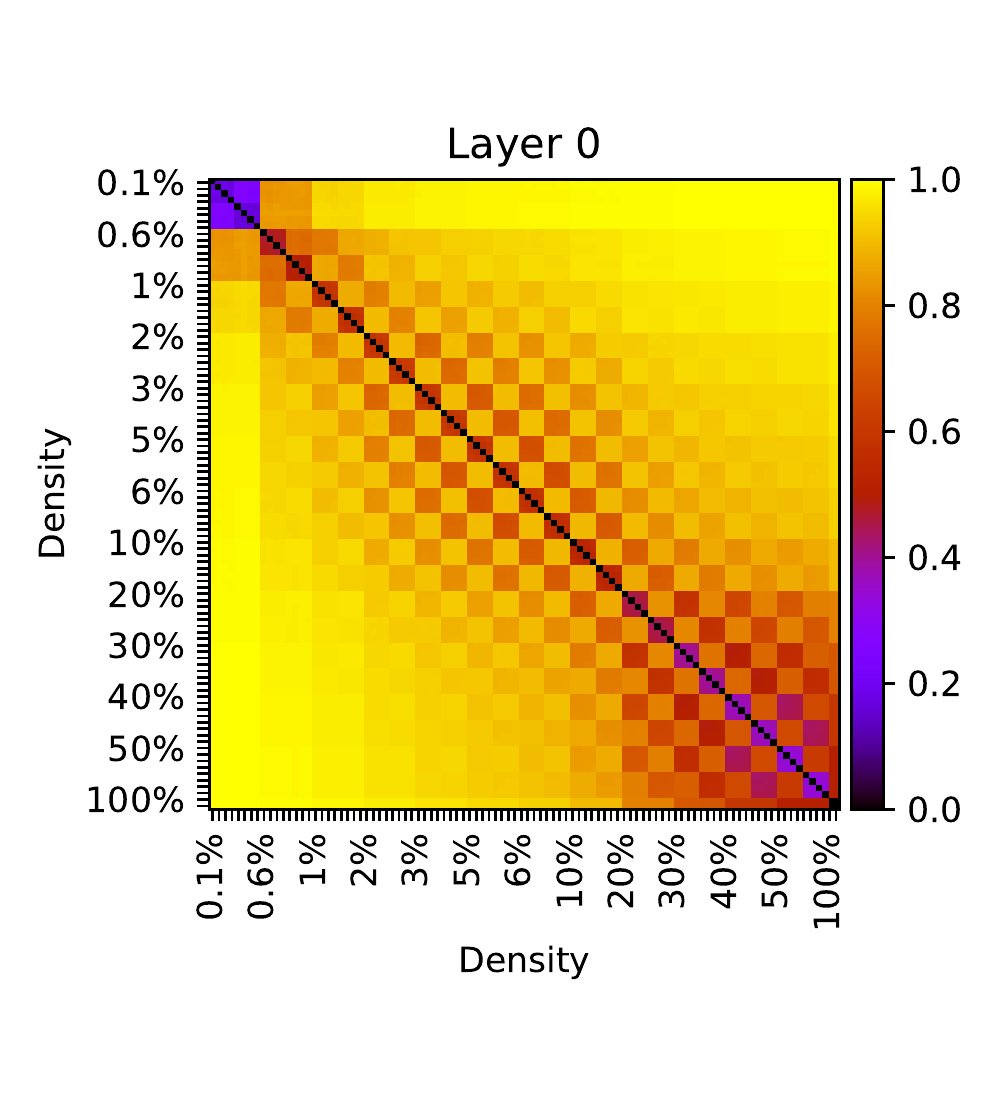}
    \caption{The final distance.}
    \label{fig:evolve_96_final}
\end{subfigure}
 \hspace*{-1.2cm} 
~
\caption{Heatmap representing the topological distance between the first layer of the 96 three-layers SET-MLP networks on Fashion-MNIST.}
\label{fig:evolve_96}
\end{figure}

First, we want to study that, initialized with very similar structures, how the topologies of these networks change when they are optimized by adaptive sparse connectivity. We choose the same 10 networks generated for Fashion-MNIST in Figure \ref{fig:only5} and train them with SET for 200 epochs. All the hyper-parameters are the same as in Section \ref{para:performance}. We apply NNSTD to measure the pairwise topological distance among these 10 networks at the $10^{th}$, the $30^{th}$, the  $100^{th}$ and the $190^{th}$ epoch. 
\begin{table}[ht]
\caption{The test accuracy of networks used for the evolutionary optimization process of adaptive sparse connectivity in Section \ref{evolution}, in percentage.}
\label{table_trials}
\begin{center}
\begin{tabular}{l|cccccccccc}
\toprule
 & W0 & W1 & W2 & W3 & W4 & W5 & W6 & W7 & W8 & W9 \\
\midrule 
Fashion-MNIST & 87.48 & 87.53 & 87.41 & 87.54 & 88.01 & 87.58 & 87.34 & 87.70 & 87.77 & 88.02 \\
CIFAR-10 & 65.46 & 65.62 & 65.26 & 65.46 & 65.00 & 65.57 & 65.61 & 64.92 & 64.86 & 65.58 \\
\bottomrule
\end{tabular}
\end{center}
\end{table}
It can be observed in Figure \ref{fig:e5} that, while initialized similarly, the topological distance between networks gradually increases from 0 to 0.6. This means that similar initial topologies gradually evolve to very different topologies while training with adaptive sparse connectivity. It is worth noting that while these networks end up with very different topologies, they achieve very similar test accuracy, as shown in Table \ref{table_trials}. This phenomenon shows that there are many sparse topologies obtained by adaptive sparse connectivity that can achieve good performance. This result can be treated as the complement of \textit{Lottery Ticket Hypothesis}, which claims that, with ``lucky'' initialization, there are subnetworks yielding an equal or even better test accuracy than the original network. We empirically demonstrate that many sub-networks having good performance can be found by adaptive sparse connectivity, even without the ``lucky'' initialization. Besides, Figure \ref{fig:evolve_96} depicts the comparison between the initial and the final topological distance among the 96 networks used in Figure \ref{fig:tc1}. We can see that the distance among different networks also increases after the training process in varying degrees.

Second, we conduct a controlled experiment to study the evolutionary trajectory of networks with very different topologies. We train 10 two-layer SET-MLPs on CIFAR-10 for 900 epochs. All the hyperparameters of these 10 networks are the same except for random seeds. The density level that we choose for this experiment is 0.7\%. With this setup, all the networks have very different topologies even with the same density level. The topologies are optimized by adaptive sparse connectivity (prune-and-regrow strategy) during training with a pruning rate of 20\% and the weights are optimized by momentum SGD with a learning rate of 0.01. 

\begin{figure}[!ht]
\centering
\begin{subfigure}[b]{0.30\textwidth}
    \includegraphics[width=\textwidth]{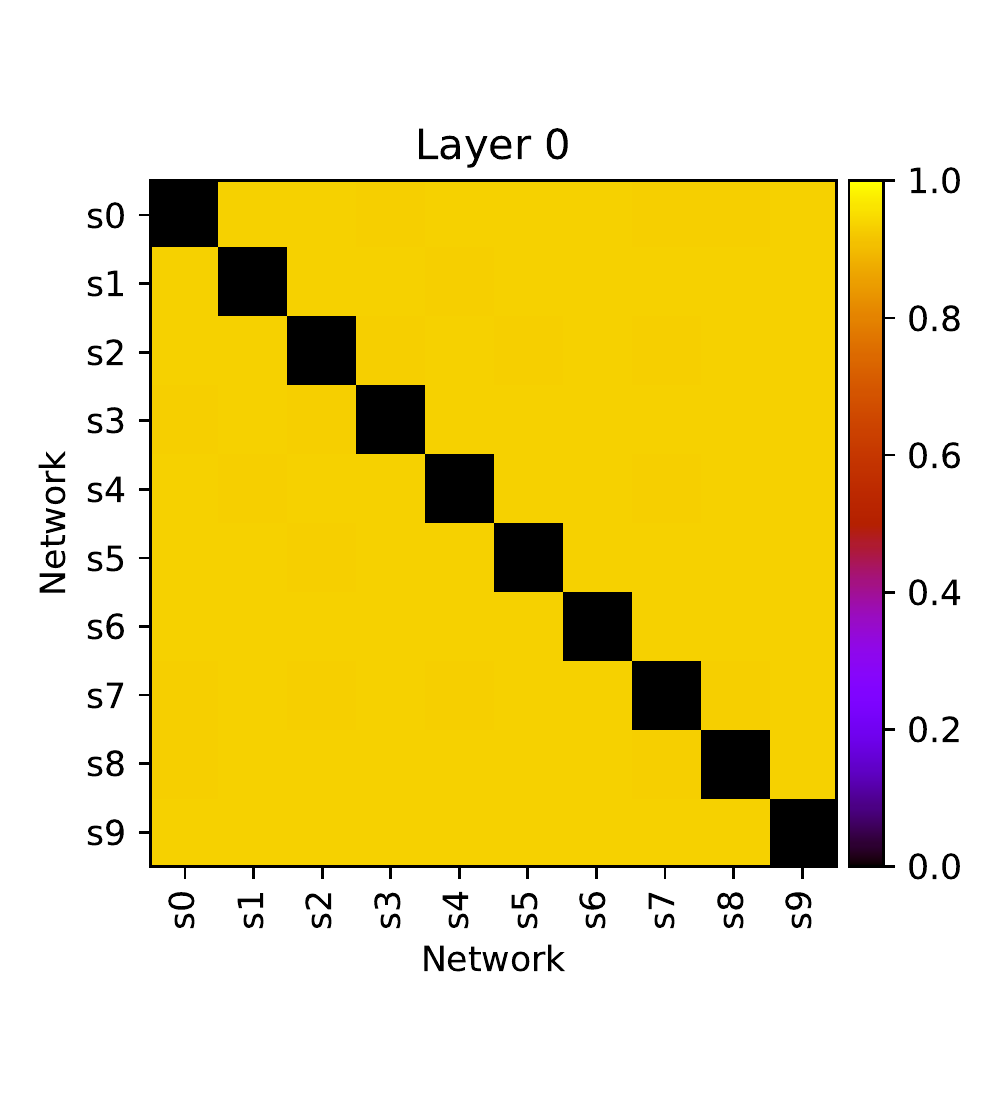}
     \caption{}
    \label{fig:hm1}
\end{subfigure}
~
\begin{subfigure}[b]{0.30\textwidth}
    \includegraphics[width=\textwidth]{heatmapf_l0compare_paths_edit_opti_SET_cf10_e5_initial_epoch000_cf10}
    \caption{}
    \label{fig:hm2}
\end{subfigure}
~
\begin{subfigure}[b]{0.28\textwidth}
    \includegraphics[width=1.07\textwidth]{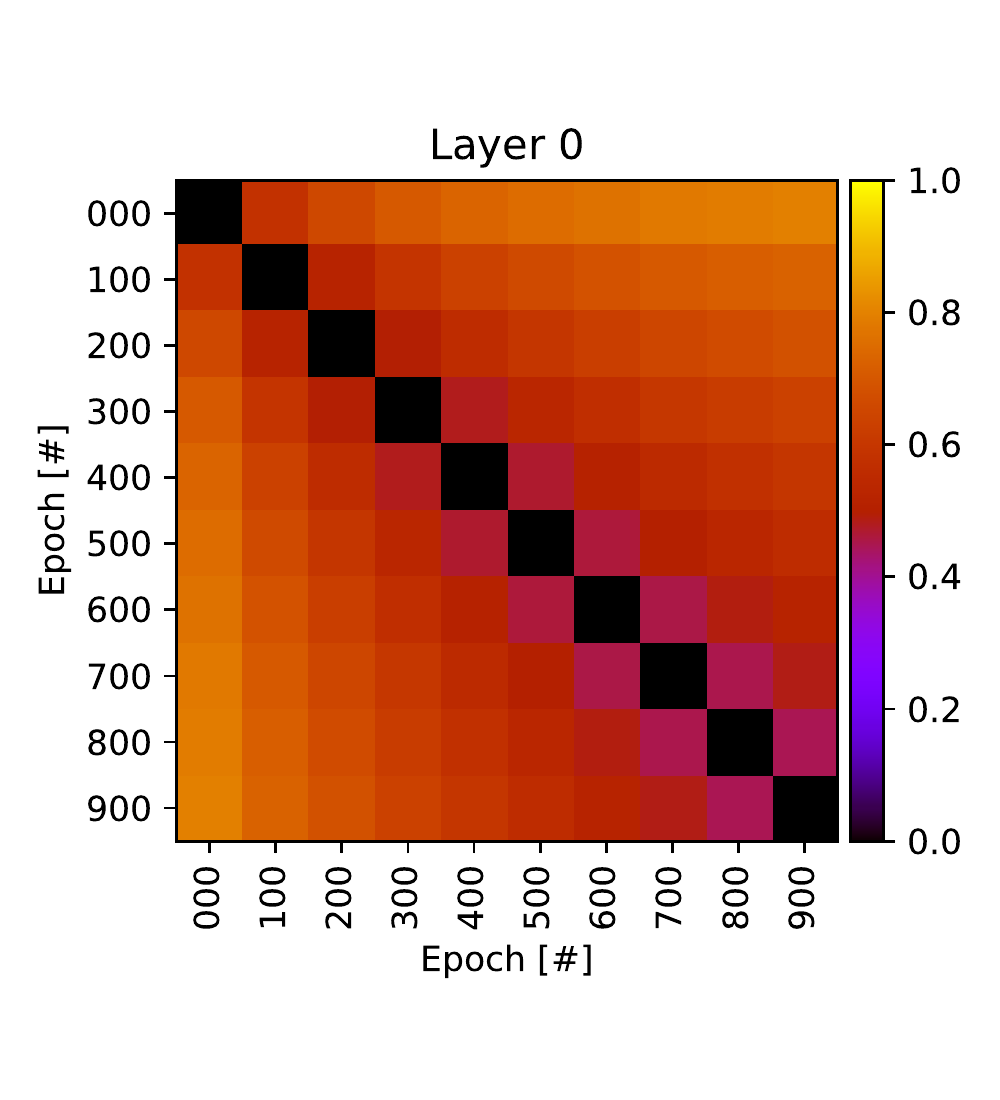}
    \caption{}
    \label{fig:hm3}
\end{subfigure}
\caption{Heatmap representing the topological distance between the first layer of the two-layer SET-MLP networks on CIFAR-10. (a) refers to distance before training. (b) refers to distance after training. (c) represents the topological distance evolution during training for the first network. $s_i (i=1,2,...,9)$ represents different networks.}
\label{fig:hm}
\end{figure}

The distance between different networks before training is very big as they are generated with different random seeds (Figure \ref{fig:hm1}), while the expectation is that these networks will end up after the training process also with very different topologies. This is clearly reflected in Figure \ref{fig:hm2}. 

We are also interested in how the topology evolves within one network trained with SET. Are the difference between the final topology and the original topology big or small? To answer this question, we visualize the optimization process of the sparse topology during training within one network. We save the topologies obtained every 100 epochs and we use the proposed method to compare them with each other. The result is illustrated in Figure \ref{fig:hm3}. We can see that the topological distance gradually increases from 0 to a big value, around 0.8. This means that, initialized with a random sparse topology, the network evolves towards a totally different topology during training.

In all cases, after training, the topologies end up with very different sparse configurations, while at the same time all of them have very similar performance as shown in Table \ref{table_trials}. We highlight that this phenomenon is in line with Fashion-MNIST, which confirms our observation that there is a plenitude of sparse topologies obtained by adaptive sparse connectivity which achieve very good performance.

\subsection{Combinatorial Optimization of Sparse Neural Networks}
\label{para:verifyASC}
\begin{figure}[!ht]
\centering
\includegraphics[width=0.6\textwidth]{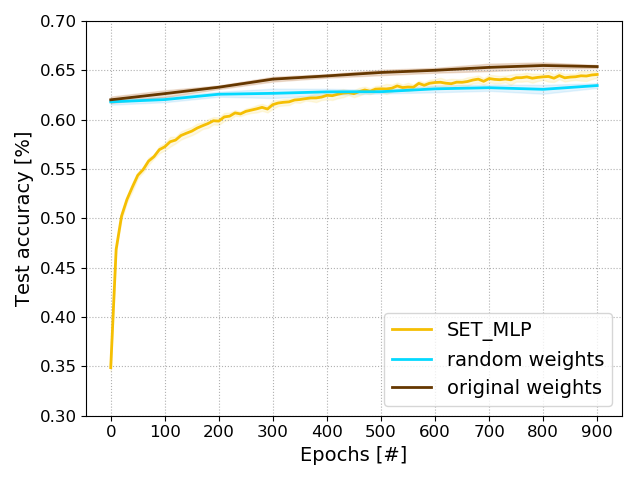}
\caption{Average test accuracy convergence of the SET network (yellow) and the average test accuracy of the retrained networks:  starting from SET weights values (brown) and starting from random weights values (vivid cyan). Each line is the average of 10 trials.}
\label{fig:retrain_cifar10}
\end{figure}
Although the sparse networks with fixed topology are able to reach similar performance with dense models, randomly initialized sparse networks can not always guarantee good performance, especially when the sparsity is very high as shown in Figure \ref{fig:Accuracy_COM}. One effective way to optimize the sparse topology is adaptive sparse connectivity, a technique based on connection pruning followed by connection regrowing, which has shown good performance in the previous works \cite{mocanu2018scalable,mostafa2019parameter,dettmers2019sparse,evci2019rigging}. Essentially, the learning process of the above-mentioned techniques based on adaptive sparse connectivity is a combinatorial optimization problem (model parameters and sparse topologies). The good performance achieved by these techniques can not be solely achieved by the sparse topologies, nor by their initialization \cite{mostafa2019parameter}. 

Here, we want to further analyze if the topologies optimized by adaptive sparse connectivity contribute to better test accuracy or not. We hypothesize that, the test accuracy of the optimized topologies should continuously be improving until they converge. To test our hypothesis, we first initialize 10 two-layer MLPs with an extremely low density level (0.5\%) under different random seeds and then train them using SET with a pruning rate of 0.2 for 900 epochs on CIFAR-10. We save the sparse networks per 100 epochs and retrain these networks for another 1000 epochs with randomly re-initialized weights. Besides this, to sanity check the effectiveness of the combinatorial optimization, we also retrain the saved networks for 1000 epochs starting from the learned weights by SET. 

Figure \ref{fig:retrain_cifar10} plots the learning curves of SET and the averaged test accuracy of the retrained networks. We can observe that, the test accuracy of random initialized networks consistently increases as the training epoch increases. This behavior highlights the fact that the adaptive sparse connectivity indeed helps the sparse topology to evolve towards an optimal one. Besides this, it seems that the topology learns faster at the beginning. However, the retrained networks which start from random initialized weights no longer match the performance of SET after about 400 epochs, which indicates that both, the weight optimization and the topology optimization, are crucial to the performance of sparse neural networks. Compared with random re-initialization, training further with the original weights is able to significantly improve the performance.  This phenomenon provides a good indication on the behavior of sparse neural networks. It may also pinpoint directions for future research on sparse neural connectivity optimization, which, however, is out of the scope of this paper.
\section{Conclusion}
In this work, we propose the first method which can compare different sparse neural network topologies, namely NNSTD, based on graph theory. Using this method, we obtain novel insights into sparse neural networks by visualizing the topological optimization process of Sparse Evolutionary Training (SET). We demonstrate that random initialized sparse neural networks can be a good choice to substitute over-parameterized dense networks when there are no particularly high requirements for accuracy. Additionally, we show that there are many low-dimensional structures (sparse neural networks) that always achieve very good accuracy (better than dense networks) and adaptive sparse connectivity is an effective technique to find them.

In the light of these new insights, we suggest that, instead of exploring all resources to train over-parameterized models, intrinsically sparse networks with topological optimizers can be an alternative approach, as our results demonstrate that randomly initialized sparse neural networks with adaptive sparse connectivity offer benefits not just in terms of computational and memory costs, but also in terms of the principal performance criteria for neural networks, e.g. accuracy for classification tasks.

In the future, we intend to investigate larger datasets, like Imagenet \cite{russakovsky2015imagenet}, while considering also other types of sparse neural networks and other techniques to train sparse networks from scratch. We intend to invest more in developing hardware-friendly methods to induce sparsity. 
\section*{Acknowledgements}
This research has been partly funded by the NWO EDIC project.

%
%
\bibliographystyle{splncs04}
\bibliography{bibliography}

\begin{thebibliography}{10}
\providecommand{\url}[1]{\texttt{#1}}
\providecommand{\urlprefix}{URL }
\providecommand{\doi}[1]{https://doi.org/#1}

\bibitem{chauvin1989back}
Chauvin, Y.: A back-propagation algorithm with optimal use of hidden units. In:
  Advances in neural information processing systems. pp. 519--526 (1989)

\bibitem{dai2018grow}
Dai, X., Yin, H., Jha, N.K.: Grow and prune compact, fast, and accurate lstms.
  arXiv preprint arXiv:1805.11797  (2018)

\bibitem{dettmers2019sparse}
Dettmers, T., Zettlemoyer, L.: Sparse networks from scratch: Faster training
  without losing performance. arXiv preprint arXiv:1907.04840  (2019)

\bibitem{devlin2018bert}
Devlin, J., Chang, M.W., Lee, K., Toutanova, K.: Bert: Pre-training of deep
  bidirectional transformers for language understanding. arXiv preprint
  arXiv:1810.04805  (2018)

\bibitem{evci2019rigging}
Evci, U., Gale, T., Menick, J., Castro, P.S., Elsen, E.: Rigging the lottery:
  Making all tickets winners. arXiv preprint arXiv:1911.11134  (2019)

\bibitem{frankle2018lottery}
Frankle, J., Carbin, M.: The lottery ticket hypothesis: Finding sparse,
  trainable neural networks. arXiv preprint arXiv:1803.03635  (2018)

\bibitem{gale2019state}
Gale, T., Elsen, E., Hooker, S.: The state of sparsity in deep neural networks.
  arXiv preprint arXiv:1902.09574  (2019)

\bibitem{gilbert1959random}
Gilbert, E.N.: Random graphs. The Annals of Mathematical Statistics
  \textbf{30}(4),  1141--1144 (1959)

\bibitem{guo2016dynamic}
Guo, Y., Yao, A., Chen, Y.: Dynamic network surgery for efficient dnns. In:
  Advances in neural information processing systems. pp. 1379--1387 (2016)

\bibitem{han2015learning}
Han, S., Pool, J., Tran, J., Dally, W.: Learning both weights and connections
  for efficient neural network. In: Advances in neural information processing
  systems. pp. 1135--1143 (2015)

\bibitem{hassibi1993second}
Hassibi, B., Stork, D.G.: Second order derivatives for network pruning: Optimal
  brain surgeon. In: Advances in neural information processing systems. pp.
  164--171 (1993)

\bibitem{he2018soft}
He, Y., Kang, G., Dong, X., Fu, Y., Yang, Y.: Soft filter pruning for
  accelerating deep convolutional neural networks. arXiv preprint
  arXiv:1808.06866  (2018)

\bibitem{he2019filter}
He, Y., Liu, P., Wang, Z., Hu, Z., Yang, Y.: Filter pruning via geometric
  median for deep convolutional neural networks acceleration. In: Proceedings
  of the IEEE Conference on Computer Vision and Pattern Recognition. pp.
  4340--4349 (2019)

\bibitem{jin2016deep}
Jin, X., Xu, C., Feng, J., Wei, Y., Xiong, J., Yan, S.: Deep learning with
  s-shaped rectified linear activation units. In: Thirtieth AAAI Conference on
  Artificial Intelligence (2016)

\bibitem{krizhevsky2009learning}
Krizhevsky, A., Hinton, G., et~al.: Learning multiple layers of features from
  tiny images. Tech. rep., Citeseer (2009)

\bibitem{lecun1990optimal}
LeCun, Y., Denker, J.S., Solla, S.A.: Optimal brain damage. In: Advances in
  neural information processing systems. pp. 598--605 (1990)

\bibitem{lee2018snip}
Lee, N., Ajanthan, T., Torr, P.H.: Snip: Single-shot network pruning based on
  connection sensitivity. arXiv preprint arXiv:1810.02340  (2018)

\bibitem{li2015convergent}
Li, Y., Yosinski, J., Clune, J., Lipson, H., Hopcroft, J.E.: Convergent
  learning: Do different neural networks learn the same representations? In:
  FE@ NIPS. pp. 196--212 (2015)

\bibitem{lin2017runtime}
Lin, J., Rao, Y., Lu, J., Zhou, J.: Runtime neural pruning. In: Advances in
  Neural Information Processing Systems. pp. 2181--2191 (2017)

\bibitem{liu2019intrinsically}
Liu, S., Mocanu, D.C., Pechenizkiy, M.: Intrinsically sparse long short-term
  memory networks. arXiv preprint arXiv:1901.09208  (2019)

\bibitem{liu2018rethinking}
Liu, Z., Sun, M., Zhou, T., Huang, G., Darrell, T.: Rethinking the value of
  network pruning. arXiv preprint arXiv:1810.05270  (2018)

\bibitem{louizos2017bayesian}
Louizos, C., Ullrich, K., Welling, M.: Bayesian compression for deep learning.
  In: Advances in Neural Information Processing Systems. pp. 3288--3298 (2017)

\bibitem{louizos2017learning}
Louizos, C., Welling, M., Kingma, D.P.: Learning sparse neural networks through
  $ l\_0 $ regularization. arXiv preprint arXiv:1712.01312  (2017)

\bibitem{dcmocanuphd}
Mocanu, D.C.: Network computations in artificial intelligence. Ph.D. thesis
  (2017)

\bibitem{gotMocanuEcml2016}
Mocanu, D.C., Mocanu, E., Nguyen, P.H., Gibescu, M., Liotta, A.: A topological
  insight into restricted boltzmann machines. Machine Learning
  \textbf{104}(2),  243--270 (2016). \doi{10.1007/s10994-016-5570-z}

\bibitem{mocanu2018scalable}
Mocanu, D.C., Mocanu, E., Stone, P., Nguyen, P.H., Gibescu, M., Liotta, A.:
  Scalable training of artificial neural networks with adaptive sparse
  connectivity inspired by network science. Nature communications
  \textbf{9}(1), ~2383 (2018)

\bibitem{molchanov2017variational}
Molchanov, D., Ashukha, A., Vetrov, D.: Variational dropout sparsifies deep
  neural networks. In: Proceedings of the 34th International Conference on
  Machine Learning-Volume 70. pp. 2498--2507. JMLR. org (2017)

\bibitem{mostafa2019parameter}
Mostafa, H., Wang, X.: Parameter efficient training of deep convolutional
  neural networks by dynamic sparse reparameterization. arXiv preprint
  arXiv:1902.05967  (2019)

\bibitem{narang2017exploring}
Narang, S., Elsen, E., Diamos, G., Sengupta, S.: Exploring sparsity in
  recurrent neural networks. arXiv preprint arXiv:1704.05119  (2017)

\bibitem{russakovsky2015imagenet}
Russakovsky, O., Deng, J., Su, H., Krause, J., Satheesh, S., Ma, S., Huang, Z.,
  Karpathy, A., Khosla, A., Bernstein, M., et~al.: Imagenet large scale visual
  recognition challenge. International journal of computer vision
  \textbf{115}(3),  211--252 (2015)

\bibitem{6313167}
{Sanfeliu}, A., {Fu}, K.: A distance measure between attributed relational
  graphs for pattern recognition. IEEE Transactions on Systems, Man, and
  Cybernetics  \textbf{SMC-13}(3),  353--362 (May 1983).
  \doi{10.1109/TSMC.1983.6313167}

\bibitem{wang2020picking}
Wang, C., Zhang, G., Grosse, R.: Picking winning tickets before training by
  preserving gradient flow. arXiv preprint arXiv:2002.07376  (2020)

\bibitem{xiao2017fashion}
Xiao, H., Rasul, K., Vollgraf, R.: Fashion-mnist: a novel image dataset for
  benchmarking machine learning algorithms. arXiv preprint arXiv:1708.07747
  (2017)

\bibitem{zhou2019deconstructing}
Zhou, H., Lan, J., Liu, R., Yosinski, J.: Deconstructing lottery tickets:
  Zeros, signs, and the supermask. arXiv preprint arXiv:1905.01067  (2019)

\bibitem{zhu2017prune}
Zhu, M., Gupta, S.: To prune, or not to prune: exploring the efficacy of
  pruning for model compression. arXiv preprint arXiv:1710.01878  (2017)

\end{thebibliography}




\end{document}